\documentclass[sigconf,nonacm]{acmart}

\makeatletter
\def\@ACM@checkaffil{
    \if@ACM@instpresent\else
    \ClassWarningNoLine{\@classname}{No institution present for an affiliation}%
    \fi
    \if@ACM@citypresent\else
    \ClassWarningNoLine{\@classname}{No city present for an affiliation}%
    \fi
    \if@ACM@countrypresent\else
        \ClassWarningNoLine{\@classname}{No country present for an affiliation}%
    \fi
}
\makeatother

\AtBeginDocument{%
  \providecommand\BibTeX{{%
    \normalfont B\kern-0.5em{\scshape i\kern-0.25em b}\kern-0.8em\TeX}}}

\setcopyright{acmcopyright}
\copyrightyear{2018}
\acmYear{2018}
\acmDOI{XXXXXXX.XXXXXXX}

\usepackage{float}
\usepackage{graphicx}
\usepackage{algorithm}
\usepackage{algpseudocode}
\usepackage{dblfloatfix}
\usepackage{float}
\usepackage{placeins}

\begin{document}

\title{Data Cross-Segmentation for Improved Generalization in Reinforcement Learning Based Algorithmic Trading}

\author{Vikram Duvvur*}
\affiliation{%
  \institution{Carnegie Mellon University}}
\email{vduvvur@cs.cmu.edu}

\author{Aashay Mehta*}
\affiliation{%
  \institution{Carnegie Mellon University}}
\email{aashaypm@cs.cmu.edu}

\author{Edward Sun*}
\affiliation{%
  \institution{Carnegie Mellon University}}
\email{edwardsu@cs.cmu.edu}

\author{Bo Wu*}
\affiliation{%
  \institution{Carnegie Mellon University}
  }
\email{bw1@cs.cmu.edu}

\author{Ken Yew Chan}
\affiliation{%
 \institution{Kenanga Investment Bank Berhad}
}
\email{kychan@kenanga.com.my}

\author{Jeff Schneider}
\affiliation{%
 \institution{Carnegie Mellon University}
}
\email{jeff.schneider@cs.cmu.edu}

\renewcommand{\shortauthors}{Duvvur, Mehta, Sun, Wu, Chan, Schneider}

\begin{abstract}
  The use of machine learning in algorithmic trading systems is increasingly common.
  In a typical set-up, supervised learning is used to predict the future prices of assets, and those predictions drive a simple trading and execution strategy.  This is quite effective when the predictions have sufficient signal, markets are liquid, and transaction costs are low.  However, those conditions often do not hold in thinly traded financial markets and markets for differentiated assets such as real estate or vehicles.  In these markets, the trading strategy must consider the long-term effects of taking positions that are relatively more difficult to change.  In this work, we propose a Reinforcement Learning (RL) algorithm that trades based on signals from a learned predictive model and addresses these challenges.  We test our algorithm on 20+ years of equity data from Bursa Malaysia.
\end{abstract}

\maketitle

\def\thefootnote{*}\footnotetext{These authors contributed equally to this work, ordered alphabetically}\def\thefootnote{\arabic{footnote}}

\section{Introduction}

Deep learning has revolutionized the field of artificial intelligence. In particular, it has gained widespread usage in applications such as natural language processing and computer vision \cite{brown2020language, ramesh2022hierarchical}. 

Finance is another area that has witnessed increasing attention from both industry and academia. The problem of stock price prediction has especially seen tremendous progress in recent years \cite{JIANG2021115537}. However, prediction isn't the end goal. We want to use these predictions to trade and profit. This may seem straightforward - buy a stock if predicted to go up significantly, and sell it if the opposite is true. But this requires us to make simplifying assumptions about market liquidity and transaction costs. 

In this paper, we propose an approach that combines supervised prediction with reinforcement learning for trading in illiquid markets with high friction costs. We use a deep learning model to make predictions about the movements of various stocks and an RL agent to make trading decisions based on these predictions.

Further, we demonstrate the importance of training the prediction and RL policy models on non-overlapping datasets. We go on to show that our data cross-segmentation process leads to models that generalize better.

\section{Related Work}
Reinforcement Learning is a machine learning technique that teaches agents how to act and accomplish goals in dynamic environments, including stock trading. RL is a trial-and-error process for optimal behavior discovery guided by rewards. RL in stock trading attempts to learn a policy that maximizes returns. This section highlights the contributions of previous work on applying RL to algorithmic trading.

\citet{lee2007multiagent} was one of the first to explore RL in daily stock trading. The work proposed a multiagent Q-learning based strategy, in which four cooperative Q-learning agents work together to maximize profit. The first agent is buy signal agent, which aims to find the right time to buy KOSPI. The second agent is the opposite and produces sell signals. The next two agents are the buy and sell order agents, which are responsible for intraday order execution. The paper uses a mix of technical analysis and indicators as input features and is trained on historical data to learn optimal policies.

More recently, works involving deep reinforcement learning (DRL) have shown success due to its ability to process higher dimensional market related data and incorporate that into the agent's state space. Two similar works \citet{liu2018practical} and  \citet{oshingbesan2022cmu} explore DRL algorithms applied on the multi-asset Dow Jones 30 company case. In both studies, the final or best model relied on an actor-critic framework, with one being advantage actor critic (A2C) \cite{mnih2016asynchronous} and the other being deep deterministic policy gradient (DDPG) \cite{lillicrap2015continuous}. DDPG with no transaction costs and daily portfolio return as the reward resulted in 1.79 sharpe \cite{liu2018practical}. A2C at 0, 0.1\%, and 1\% transaction costs with log return and sharpe ratio rewards remained consistent around 1 sharpe for all combinations \cite{oshingbesan2022cmu}.

\begin{figure*}[t!]
    \centering
    \includegraphics[width=\textwidth]{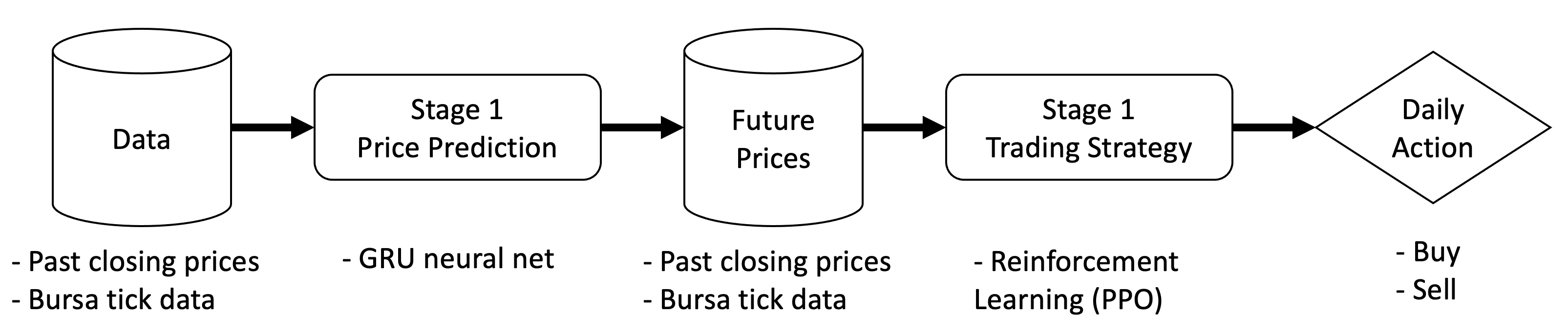}
    \caption{Model Pipeline}
    \label{fig:pipeline}
\end{figure*}

\citet{zhang2020cost} designed a framework to manage a portfolio of multiple assets simultaneously and trained a policy to make portfolio rebalancing decisions while factoring cross asset correlation and transaction costs. Their approach involves: reward engineering to penalize empirical log return variance plus transaction costs incurred after each action, sequential feature mapping through LSTMs, and correlation mapping using simple convolutions. Lastly, RL training is done through DDPG, which is effective for continuous action spaces such as asset allocation weighting. One of this paper's strengths is the cost-sensitive reward, which is more representative of what traders consider in real-world stock trading.

Another automated portfolio trading system is by \citet{li2021auto}, who proposes a trading system that contains a feature preprocessing module and a trading module. The feature preprocessing module transforms technical indicator data through principal component analysis (PCA) and discrete wavelet transform (DWT). The output is sent to the trading module, which is multiplied by some learned weight matrix and passed through non-linear transforms to compute portfolio returns during training. Through Recurrent RL (RRL), this weight matrix is optimized by gradient ascent with respect to sharpe ratio. Their results look promising, with the majority slightly above 1.00 sharpe at 0.1\% transaction costs.

\citet{meng2019reinforcement} reviewed RL techniques applied to financial markets, including, genetic algorithms, recurrent RL, deep Q-learning, and on/off-policy RL. \citet{sun2021rlqt} surveyed research progress on RL-based quantitative trading tasks. They grouped tasks into four categories: algorithmic trading, portfolio management, order execution, and market making. The paper then detailed differences between value-based, policy-based, actor-critic, and model-based RL algorithms across these major categories. It also noted a recent trend of DRL outperforming traditional RL methods.

The application of RL to algorithmic trading is an active field of research, with an ongoing effort to design novel trading strategies through a mix of expertise from finance and machine learning backgrounds. New RL techniques like inverse RL are constantly implemented and tested in trading environments \cite{halperin2022inverse}. However, a common limiting factor to all such approaches is their predictive ability. Even the most optimal RL policy cannot learn to generate profit if their price forecasting accuracy is only 50\%. Understandably, this is one of the most heavily invested in research areas due to core impact it has on all market participants. 

\section{Methods}

We employ a two-stage pipeline to tackle our task. The first stage is a prediction stage, where a model is trained to predict stock price movement based on the given data. This produces secondary data of predicted stock price movements. Based on this new data, we train a reinforcement learning model in the second stage to learn actual trading strategies. This split structure allows us to use specialized models and different learning strategies for each task. It also allows us to flexibly interchange each component should there arise any independent improvements to any of the components. For example, a company may already have a strong undisclosed model for predicting stock movement. They are then able to drop it in for an enhanced performance without modifying any other parts of the pipeline.

\subsection{Data}

The data is based on stock information that can be obtained for any stock market. Our data is on the Malaysian stock market, courtesy of Kenanga Bank of Malaysia. A 30-day history of the following features are used, each of which are either raw data or computed from raw data.
\begin{itemize}
    \item closing price
    \item moving averages
    \item realized volatility
    \item relative strength index
\end{itemize}
We document the full set of features and their computation in Appendix \ref{appendix:a}. The label is the closing price of the next day. For consistency, we normalize each feature to zero mean and unit standard deviation. We normalize the label using the same normalization parameters as the closing price feature.

This defines a single input data point $X_0$ as a $D\times H$ matrix where $D$ is the number of days to include in the history and $H$ is the number of features.

\subsection{Price prediction model}

We use a GRU (gated recurrent unit) neural network to perform price prediction. The choice was due to its simplicity and recurrence, which is applicable to time-series data.

\subsubsection{GRU network}

The GRU network transforms the input as follows.
\begin{enumerate}
    \item Linear layer: $X_1=\text{ReLU}(\text{BatchNorm}(X_0W_1+ b_1))$
    \item GRU layer: $X_2=\text{BatchNorm}(\text{GRU}(X_1))$ where GRU uses the default implementation by PyTorch. This notation compresses the recurrence along the time-axis into a single operation, since we feed $D$ days of data at the same time.
    \item Weighing linear layer: $A=\text{Softmax}(\text{tanh}(\text{BatchNorm}(X_2W_3 + b_3)))$
    \item Weighing: $X_3 = X_2A$
    \item Final layer: $\widehat{y}=X_3W_4 + b_4$
\end{enumerate}
In our experiments we set sizes of $W$ and $b$ such that $X_1$ has dimensions $D\times256$, $X_2$ has dimensions $D\times512$, $A$ has dimensions $512\times512$, and the final $\widehat{y}$ is a scalar. After the prediction, we reverse the normalization on $\widehat{y}$ to recover the predicted closing price of the stock on the next day. We then finally discretize the output to 1 if the predicted price is higher than the previous closing price, and 0 otherwise.

\begin{figure*}[t!]
    \centering
    \includegraphics[width=\linewidth]{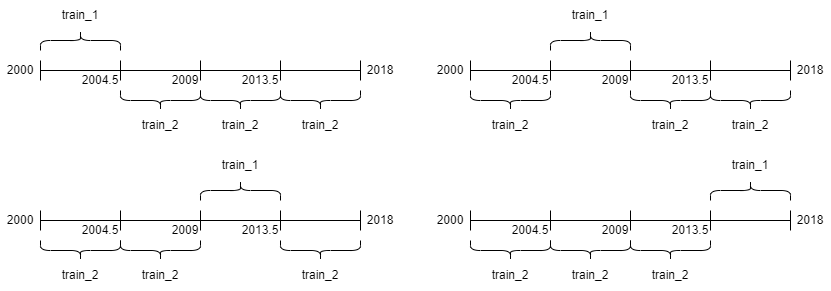}
    \caption{Illustration of splitting by quarters, yielding 12 possible cross-segmentation options.  Each train\_1 segment can be separately paired with each of the other 3 train\_2 options.}
    \label{fig:segmentation2}
\end{figure*}

\subsubsection{Ensemble}

In practice, we find that due to the noisiness of financial data, an ensemble of models has a much higher stability than a single model. In addition, the set of predicted prices from an ensemble of models can be used as a proxy for the variance of the prediction. Therefore, 
we train $M$ duplicate models with random weight initializations. Our final output is the average of the $M$ binary discretized single model outputs, which is a single floating point value in the range of 0 to 1.

\subsection{Strategy reinforcement learning}

Given the predicted stock price movements, we then use reinforcement learning to come up with a trading strategy that will consider trading costs and the long-term implications of taking and existing positions.

\subsubsection{Environment setup}

We mimic an OpenAI Gym environment \cite{openAIgym} where we define the state space, action space, and transition function. Then we are able to use external libraries to simplify the training process. 

We define the state to be a length-$N$ vector, where $N$ is the number of stocks to include in the portfolio, and each entry is the output given by the ensemble model above. Optionally, we can concatenate the current portfolio to the state space as another length-$N$ vector, where each entry is the fraction of funds currently allocated to each stock.
An action is defined as a length $N$ normalized vector (where the elements sum up to 1) that denotes the fractions of our wealth we want to invest in each stock.
The reward is defined as the percentage change in our actualized wealth, and we impose a penalty for negative returns by multiplying it by a scalar greater than one. This helps prevent the agent from overly making trades that are too risky, and controls the Sharpe ratio.

We also optionally add transaction costs to the environment, implemented simply as a percentage of the transaction value of a buy or sell.

\subsubsection{Model}

We use the PPO algorithm, since this showed the most stability among all general-purpose RL algorithms we tested. We used the implementation from stable-baselines3 \cite{stable-baselines3}.

\subsection{Data cross-segmentation}

A major problem is the lack of training data for this task. Whereas even MNIST \cite{lecun2010mnist} has 60,000 training examples, twenty years of daily closing prices only give roughly 7,300 datapoints for us to train on. Here we can employ something similar to cross-validation to enhance our performance.

Since both stages need to be trained, the data we have at hand must be used to train both of the stages. The first prediction stage is a straightforward supervised learning problem. The second stage is a reinforcement learning problem, where we need to curate an environment based on the intermediate data of predicted price movements.

In a traditional setting, one would use the original data to train the first stage, and then the intermediate data generated by the first stage to train the second stage. However, in an environment where overfitting is likely, this can cause problems. Assume that the first stage overfits. Then the intermediate data generated from the train set will be more accurate than the intermediate data generated from a test set, and hence cause a distribution shift. This can cause the second stage to overfit to the accurate data generated from the train set, and be unable to generalize to new data which is less accurate.

To remedy this, we propose a scheme in which the prediction and the RL part are trained on disjoint data. To elaborate, we first obtain two disjoint datasets, which we call train\_1 and train\_2. Then we train the first stage (price predictor) using subset train\_1. We then use the trained stage 1 model to predict prices for the train\_2 dataset into an intermediate dataset int\_train\_2. Finally, we use int\_train\_2 to train the second stage (trading agent).

Note that int\_train\_2 is generated from a dataset completely independent of the dataset used to train the first stage, so it is not the result of overfitting and has the same distribution as that of test-time. We would expect this to generalize better than a naive approach.

Since train\_1 and train\_2 only have to be independent of each other, and they train different parts of the pipeline, we can shuffle around different parts of the data to be train\_1 and train\_2, so long as they are separate. In our temporally-stratified data, if there are $k$ groups of data, then we can pick train\_1 $k$ ways and  then pick train\_2 $k-1$ ways for a total of $k(k-1)$ possible combinations. Since each group has roughly $1/k$ of the total data, we obtain an apparent $O(k)$ fold increase in data. An illustration of this method for the case of $k=4$ is shown in Fig. \ref{fig:segmentation2}, where there are 4 ways to take train\_1 and 3 ways to take train\_2 for each train\_1, resulting in 12 possible ways to train the models. Our method is summarized below:

\begin{algorithm}
\caption{Data cross-segmentation}
\begin{algorithmic}[1]
    \item Divide train data $\mathcal{D}$ into $k$ disjoint subsets $\mathcal{D}_1, \ldots, \mathcal{D}_k$.
    \item For each subset $\mathcal{D}_i$, train the price predictor on it and predict prices on the remaining data $\mathcal{D}_{-i}$  to obtain the corresponding intermediate data $\mathcal{U}_i$.
    \item Use the consequent $\mathcal{U}_1, \ldots, \mathcal{U}_k$ to train the trading agent.
\end{algorithmic}
\end{algorithm}

\begin{figure*}[h!]
   \begin{minipage}{0.5\textwidth}
     \centering
     \includegraphics[width=\linewidth]{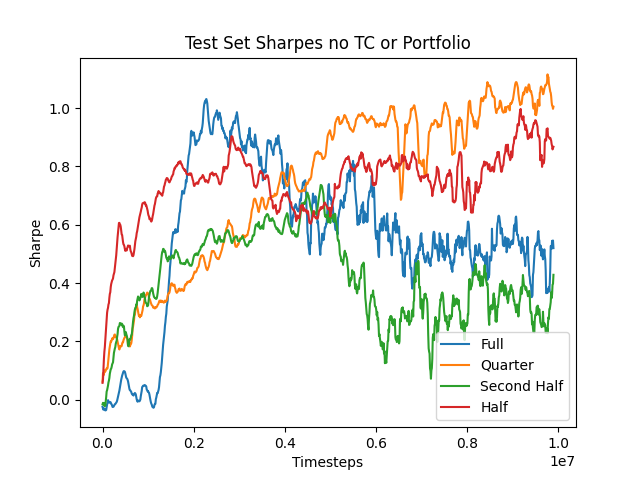}
   \end{minipage}\hfill
   \begin{minipage}{0.5\textwidth}
     \centering
     \includegraphics[width=\linewidth]{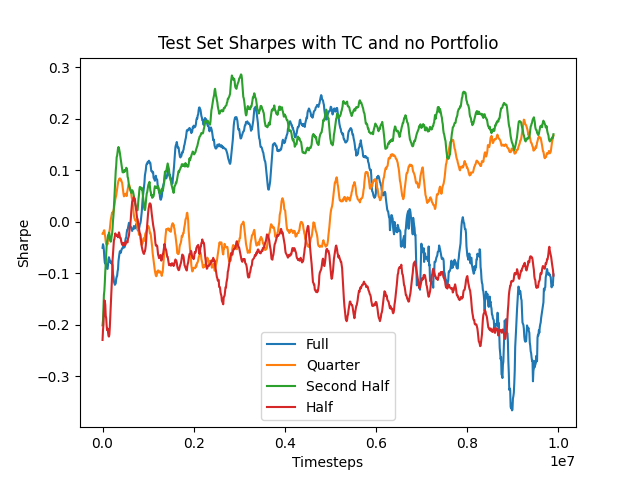}
   \end{minipage} \\
   \begin{minipage}{0.5\textwidth}
     \centering
     \includegraphics[width=\linewidth]{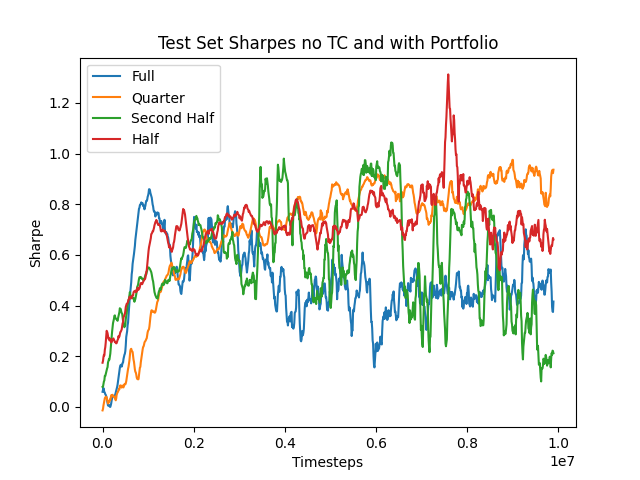}
   \end{minipage}\hfill
   \begin{minipage}{0.5\textwidth}
     \centering
     \includegraphics[width=\linewidth]{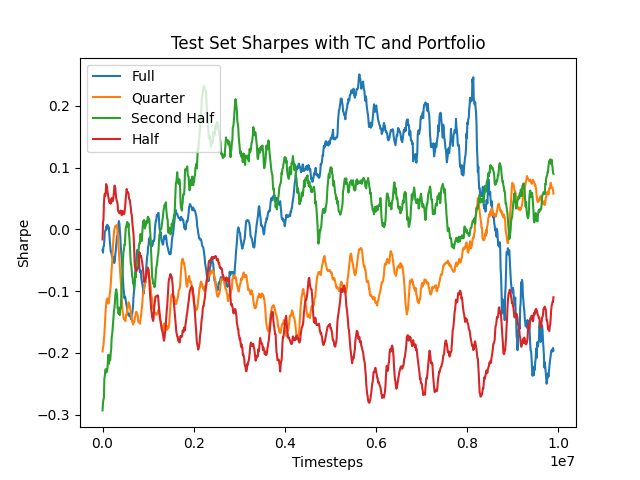}
   \end{minipage} 
    \caption{Test Sharpes over the training process for different environments}
    \label{fig:gru_test_perf}
\end{figure*}

\section{Results}

\subsection{Experimental Setup}

We partitioned our data into train, validation, and test datasets. The train data was from January 1st, 2000 to December 31st, 2018. The validation data was from January 1st, 2018 to December 31st, 2019. The test data was from January 1st, 2020 to July 20th, 2021 which was the most recent date we collected data. 

These experiments were performed on four different data splits. "Quarters" and "Halves" used data cross-segmentation to split the training data into quarters and halves, respectively. "Full" used the entire dataset to train the prediction and RL model. "Second Half" used the first half of the dataset to train the predictor, and the second half to train the RL model.

We determine the best hyperparameters for every combination of data split method (full, quarters, halves, and second half), trading cost (0\% and 0.1\%), and portfolio addition to the state space using the Optuna library \cite{optuna_2019} in Python to perform a Bayesian search. For our search, we used 50 iterations, where we average over 20 different realizations in each of them to reduce the variance of our estimate. We used the maximum of the moving average of the Sharpe ratio as our evaluation metric and trained each model for 2.5 million timesteps. Our best hyperparameter settings for the model can be found in Appendix \ref{appendix:b}.

After tuning our hyperparameters, we evaluated them on the test set with 20 random seeds and increased the number of timesteps to 10 million.

To minimize the effect of noise, all reported Sharpe values in the hyperparameter search tables and results section represent the average of the past 10 evaluation periods.

\subsection{Performance}

\begin{table}[]
    \centering
    \begin{tabular}{c|c |c}
        Threshold & Test Sharpe (no TC) & Test Sharpe (w/ TC) \\
        \hline
            0.0 & -0.04 & -0.05 \\
            0.1 & 0.34 & 0.15\\
            0.2 & 0.48 & 0.15\\
            0.3 & 0.77 & 0.35\\
            0.4 & 1.30 & 0.76\\
            0.5 & 0.74 & 0.10\\
            0.6 & 0.69 & -0.06\\
            0.7 & 1.11 & 0.21\\
            0.8 & 0.68 & -0.20\\
            0.9 & 0.83 & -0.21\\
            1.0 & 1.79 & 0.36
    \end{tabular}
    \caption{Baseline (No RL) results for different price predictor thresholds excluding and including trading costs.  Note that a threshold of 0.0 corresponds to a buy and hold strategy.}
    \label{tab:gru_baseline}
\end{table}

Across the four environment, using the full dataset to train the price predictor and trading agent is the worst option, with its test Sharpe dropping as the training process continues. Quarter cross-segmentation places in the top 2 data split methods for test set performance every time.

The overall best performing environment is the quarter cross-segmentation with no portfolio or trading costs and reaches a test Sharpe ratio of around 1. Among the experiments with trading costs, only quarter data split and second half ever ended with positive test Sharpe ratios.

When comparing to the baseline results in 
Table \ref{tab:gru_baseline}, we see that our model beats the buy and hold strategy, both with and without trading costs. It also beats many, but not all, of the threshold baseline strategies.

\subsection{Generalization}

\begin{figure*}[h!]
   \begin{minipage}{0.5\textwidth}
     \centering
     \includegraphics[width=\linewidth]{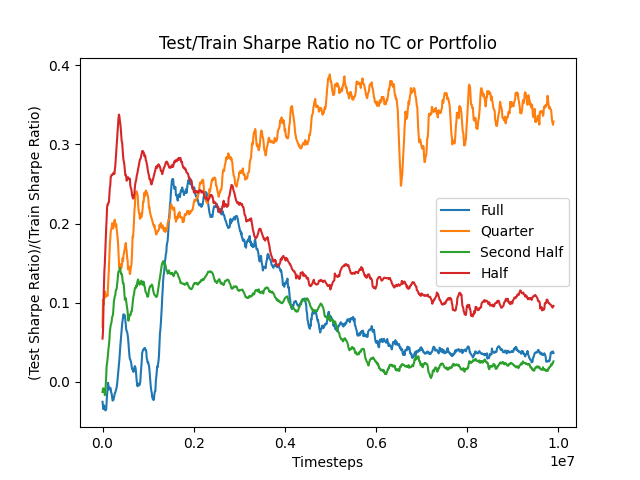}
   \end{minipage}\hfill
   \begin{minipage}{0.5\textwidth}
     \centering
     \includegraphics[width=\linewidth]{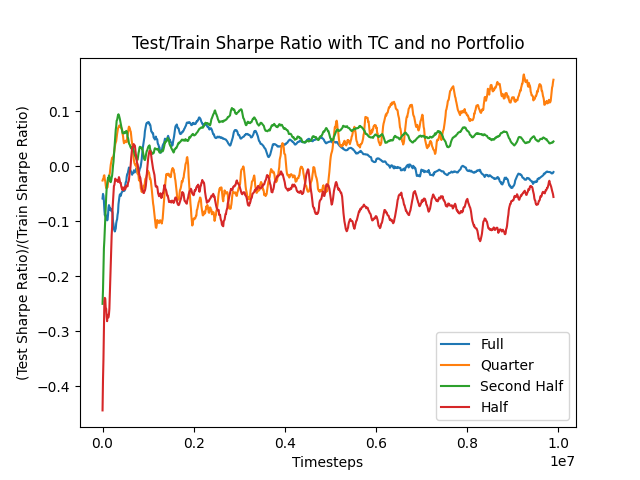}
   \end{minipage} \\
   \begin{minipage}{0.5\textwidth}
     \centering
     \includegraphics[width=\linewidth]{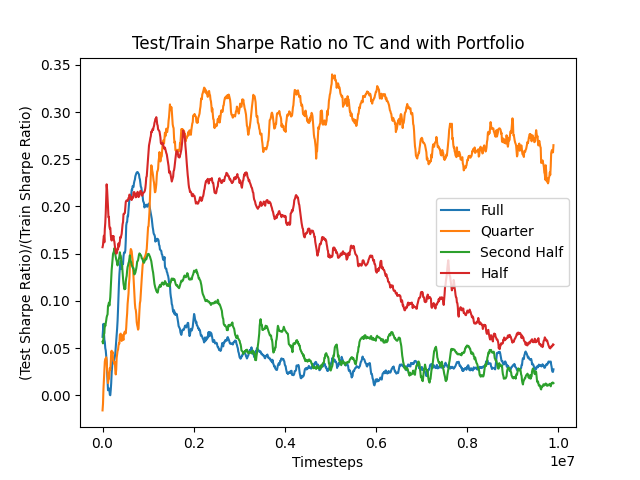}
   \end{minipage}\hfill
   \begin{minipage}{0.5\textwidth}
     \centering
     \includegraphics[width=\linewidth]{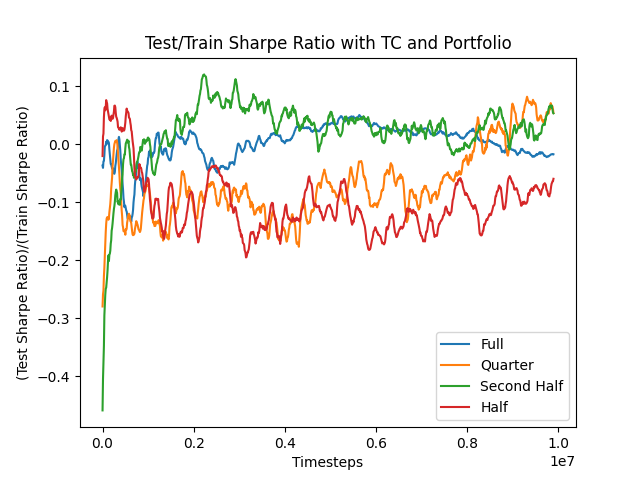}
   \end{minipage} 
    \caption{Generalization ratios over the training process for different environments}
    \label{fig:gru_test_gen}
\end{figure*}

Our metric for determining how well a model generalizes is dividing the test Sharpe of the model by the train Sharpe at that timestep. A model with perfect generalization would have a ratio of one throughout the entire training process. While independently this may not provide a lot of information, when paired with model performance we have a much easier time determining which model is the "best". 

The quarter cross-segmentation runs have the best final generalization score throughout all the environments, although the second half data split has approximately the same final generalization score as shown in Figure \ref{fig:gru_test_gen}. Noticeably, using the full data split is always the second worst in terms of generalization score, only outperforming second half. This is expected as second half has only half the data to train the RL agent on, so it can overfit more easily.

\section{Discussion}

Our results show that a RL algorithm can be used to train trading policies based on predictive models for cases that do or do not have significant trading costs.  Furthermore, we have shown that data cross-segmentation can be used to improve the performance of the trading algorithms. In particular, we see that the full algorithm that does not use data cross-segmentation overfits as the training proceeds (i.e. the test set performance in our figures declines by the end). In contrast, the performance of the training curves for the quarter algorithm increase relatively steadily through the training period.

There are many areas of potential future work that could further improve the performance of these algorithms.  One is that the current predictive model only makes 1-day predictions.  An algorithm that is considering long-term effects would likely benefit from receiving predictive signals over many time horizons (e.g. 1-day, 1-week, 1-month).  Another area is that our preliminary results indicate that models other than GRUs may deliver better signals for RL.  Finally, the actor and critic networks in the RL algorithm use simple shallow, fully connected layer networks.  We believe that an exploration of newer, attention-based networks for those components has the potential to achieve better trading decision-making.

\FloatBarrier

\bibliographystyle{ACM-Reference-Format}
\bibliography{draft}


\clearpage

\section*{Appendices}
\appendix

\section{Feature Engineering}
\label{appendix:a}

For each day we collect
\begin{itemize}
    \item Closing price
    \item Opening price
    \item High price
    \item Low price
    \item Trading volume
    \item \{5,10,20,30,40,50,60,100,120,180,200\}-day moving average
    \item Realized volatility over past \{3,5,10,20,50\} days
    \item RSI over {3,6,14,30} days
\end{itemize}
All of these were concatenated into a single row.

We defined realized volatility as the moving standard deviation of the log-returns. The log-returns are calculated as: $$r_i=\log(p_i)-\log(p_{i-1})$$ where $p_i$ refers to the price of the stock on the $i$-th day.


\begin{table*}[!htbp]
    \begin{tabular}{|l|l|l|l|l|l|l|l|l|l|}
    \hline
        \textbf{Data Split} & \textbf{TC} & \textbf{Portfolio} & \textbf{Best Val Sharpe} & \textbf{Best Test Sharpe} & \textbf{Training Time} & \textbf{Depth} & \textbf{Width} & \textbf{LR} & \textbf{Entropy} \\ \hline
        Full & 0 & 0 & 1.197 & 1.031 & 17:12:09 & 1 & 64 & 3.531E-05 & 1.310E-03 \\ \hline
        Full & 0 & 1 & 1.553 & 0.859 & 25:06:54 & 1 & 112 & 6.578E-05 & 4.015E-05 \\ \hline
        Full & 0.001 & 0 & 0.496 & 0.246 & 12:28:26 & 1 & 61 & 1.751E-04 & 1.297E-04 \\ \hline
        Full & 0.001 & 1 & 0.497 & 0.251 & 9:44:08 & 3 & 107 & 4.248E-05 & 5.324E-04 \\ \hline
        Quarters & 0 & 0 & 0.900 & 1.116 & 6:42:13 & 1 & 12 & 3.696E-04 & 3.467E-04 \\ \hline
        Quarters & 0 & 1 & 0.917 & 0.975 & 7:22:52 & 2 & 99 & 6.837E-05 & 2.748E-05 \\ \hline
        Quarters & 0.001 & 0 & 0.435 & 0.198 & 7:14:34 & 3 & 35 & 7.966E-05 & 6.604E-03 \\ \hline
        Quarters & 0.001 & 1 & 0.677 & 0.087 & 6:56:17 & 2 & 35 & 2.191E-04 & 6.486E-04 \\ \hline
        Halves & 0 & 0 & 0.923 & 0.997 & 11:00:08 & 1 & 127 & 1.547E-04 & 1.421E-05 \\ \hline
        Halves & 0 & 1 & 0.899 & 1.312 & 10:17:50 & 1 & 84 & 2.283E-04 & 5.184E-03 \\ \hline
        Halves & 0.001 & 0 & 0.494 & 0.047 & 5:00:42 & 2 & 13 & 5.862E-04 & 3.795E-03 \\ \hline
        Halves & 0.001 & 1 & 0.424 & 0.073 & 4:32:34 & 1 & 71 & 2.042E-04 & 6.104E-04 \\ \hline
        Second Half & 0 & 0 & 0.951 & 0.736 & 12:30:13 & 2 & 70 & 5.670E-05 & 7.305E-03 \\ \hline
        Second Half & 0 & 1 & 0.846 & 1.044 & 19:54:07 & 2 & 50 & 1.607E-04 & 8.764E-04 \\ \hline
        Second Half & 0.001 & 0 & 0.428 & 0.286 & 4:24:55 & 1 & 44 & 3.274E-04 & 3.065E-03 \\ \hline
        Second Half & 0.001 & 1 & 0.311 & 0.232 & 5:09:17 & 3 & 41 & 2.878E-03 & 3.416E-05 \\ \hline
    \end{tabular}
    \caption{The best hyperparameters chosen for each combination of data cross segmentation and problem setup}
    \label{gru_settings}
\end{table*}

\section{Hyperparameter Selection}
\label{appendix:b}

The table below summarizes the results of the hyperparameter tuning we did.  The hyperparameters were tuned on the validation set separately for each data split, cost, and portfolio alternative.  The best hyperparameters are shown for each combination, along with the validation Sharpe ratio they achieved (i.e. the metric they were tuned on) and their test Sharpe ratios (i.e. performance on the data they were not tuned on).

\end{document}